# Deep Transfer Network with Joint Distribution Adaptation: A New Intelligent Fault Diagnosis Framework for Industry Application

Te Han[a, b], Chao Liu[a*], Wenguang Yang[a] and Dongxiang Jiang[a, b]
a Department of Energy and Power Engineering, Tsinghua University, Beijing 100084, China
b State Key Lab of Control and Simulation of Power System and Generation Equipment, Tsinghua University, Beijing 100084, China; cliu5@tsinghua.edu.cn (Chao Liu[a]).

*Abstract*—In recent years, an increasing popularity of deep learning model for intelligent condition monitoring and diagnosis as well as prognostics used for mechanical systems and structures has been observed. In the previous studies, however, a major assumption accepted by default, is that the training and testing data are taking from same feature distribution. Unfortunately, this assumption is mostly invalid in real application, resulting in a certain lack of applicability for the traditional diagnosis approaches. Inspired by the idea of transfer learning that leverages the knowledge learnt from rich labeled data in source domain to facilitate diagnosing a new but similar target task, a new intelligent fault diagnosis framework, i.e., deep transfer network (DTN), which generalizes deep learning model to domain adaptation scenario, is proposed in this paper. By extending the marginal distribution adaptation (MDA) to joint distribution adaptation (JDA), the proposed framework can exploit the discrimination structures associated with the labeled data in source domain to adapt the conditional distribution of unlabeled target data, and thus guarantee a more accurate distribution matching. Extensive empirical evaluations on three fault datasets validate the applicability and practicability of DTN, while achieving many state-of-the-art transfer results in terms of diverse operating conditions, fault severities and fault types.

## I. INTRODUCTION

IN modern industry, machines and equipment are developing towards the direction of high-precision, high-efficiency, more automatic and more complicated, making the breakdown or even accidents more frequent. Intelligent monitoring and fault diagnosis systems, in a broad sense, have always been key to attaining the enhancement of security and reliability of industry equipment [1]. Over the past decade, various attempts have been made to design efficient algorithms or new ways for achieving superior diagnostic performance. These studies usually merge advanced signal processing algorithms and machine learning techniques to process machine data and make diagnostic decisions intelligently, leading to impressive results in many diagnosis cases [2]-[4].

To date, according to the procedure of diagnosis framework, most of previous studies can be divided into two stages. In the first stage, the diagnosis framework mainly consists of three steps 1) data collection, 2) feature extraction and selection, and 3) fault classification (Fig. 1(a)) [5]-[8]. In this framework, a massive efforts have been devoted to manual feature extraction and selection. This process benefits from the extensive domain expertise captured by diagnosis specialist, but inevitably requires a large expenditure of labor and time. Besides, the designed features always aim at special application object, and thus have limited adaptability when facing new diagnosis issues or changing the physical characteristics of the original systems [9]. Moreover, the final decision-making resorts to pattern recognition methods, and the diagnostic performance is often sensitive to model parameters, such as the penalty factor and kernel function parameter in support vector machine (SVM), indicating the additional parameter optimization procedure need to be executed [10]. To tackle these issues, an adaptive feature learning based diagnostic framework with deep learning technology is emerging in the second stage [11]-[14]. With the aid of multi-layer nonlinear modeling scheme, this framework provides an end-to-end learning procedure from input signals to output diagnosis labels, as shown in Fig. 1(b). The training process, in which the error estimated by the upper classification layer is back-propagated to update the parameters of lower feature descriptor layers, further guarantees the co-adaptation of the whole network. In the past few years, the deep learning models, including stacked auto-encoder (SAE) [15], deep belief networks (DBN) [16] and, in particular, convolutional neural networks (CNN) [17]-[23], have gained much popularity and success in mechanical fault diagnosis issues, showing an extraordinary feature learning and fitting capacity.

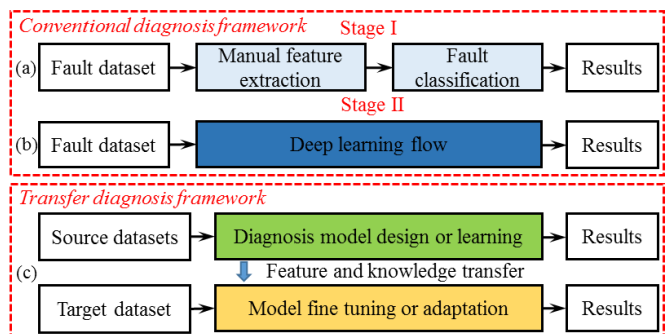

Fig. 1. Intelligent fault diagnosis framework. (a) Stage I, (b) Stage II and (c) New one.

Despite its marvelous success, the above frameworks may suffer from two latent disadvantages before extensive and flexible industry applications. 1) Most of existing studies are under the assumption of same distributions between training data and testing data, while it is not always in accord with the real situations. The diagnostic model is generally learned with the training data of limited conditions, and the generalization error cannot be large enough to guarantee the success on the testing data for diverse application domains. Thus, the diagnostic model often need to be relearned from scratch for new diagnosis tasks. 2) A large amount of training data is often required so that the hierarchical features can be fully learned and a stronger generalization ability can be achieved by the deep networks. However, in real problems, especially for those unseen conditions, collecting sufficient typically labeled samples is usually an expensive or even impossible task.

Consequently, there is, in particular, a need to develop a framework that can borrow the obtained useful information from historical tasks to facilitate diagnosing a new but similar testing task, instead of reconstructing and re-training a new diagnosis model from scratch. This could make the diagnosis systems more practical and flexible to be deployed in a variety of applications. Transfer learning provides the source of inspiration and has proven its wide applicability spanning through various fields [24]-[27]. Different from traditional machine learning procedure, transfer learning aims to improve the model performance and reduces the quantity of required sample in target domain by leveraging the transferable features or knowledge from source domain [28]. It is definitely a promising way for tackling the aforementioned challenges. On the basis of this, we propose a new intelligent fault diagnosis framework, i.e., deep transfer network (DTN), in this work, as shown in Fig. 1(c). A base network, convolutional neural networks (CNN) in this work, is first learned on sufficient fault data in source domain. Since most of real-time-monitoring data (target domain) in real industry are unlabeled, a joint distribution adaptation (JDA) scheme is devised to reduce the domain discrepancy and realize the goal of domain adaption without target supervision.

The main contributions of this work include: 1) Proposing a new intelligent fault diagnosis framework, DTN, for real industrial application, exploiting the idea of transfer learning. The DTN with JDA method can utilize the unlabeled target data to realize domain adaption, which conforms better with the real situation. 2) The diagnosis cases on three fault datasets, i.e. wind turbine dataset, bearing dataset and gearbox dataset, are conducted to explore the transfer ability of proposed method under different scenarios, namely, various operating conditions, diverse fault severity levels and different fault types. 3) The presented framework achieves superior diagnosis performance in comparison with the state-of-the-art methods including supervised and domain adaption methods. Especially, both the network visualization and convergence analysis give an intuitive presentation of adaptation results and verify effectiveness of our framework.

The remaining parts are organized as follows. In Section II, related works on CNN and transfer learning are discussed. In Section III, the proposed intelligent fault diagnosis framework, DTN with JDA, is presented. The comparison methods and implementation details are also explained. The experiments and discussion are given in Section IV. Finally, the conclusions are drawn in Section V.

## II. RELATED WORKS

### A. Convolutional neural networks

CNN, as a type of most effective deep learning models, has been widely used in image processing, computer vision and speech recognition. Typically, a CNN consists of three types of layers, which are convolutional layers, pooling layers and fully connected layers. The first step of CNN is to convolve the input signal with a set of filter kernels (1D for time-series signal and 2D for image). All the feature activations by convolution operation at different locations constitute the feature map. A nonlinear activation function, generally rectified linear unit (ReLU), is applied on the sum of feature maps. The operation of convolutional layer can be expressed as:

$$c_n^r = ReLU(\sum_m v_m^{r-1} * w_n^r + b_n^r) \quad (1)$$

where $c_n^r$ is the $nth$ output of convolutional layer $r$, $n$ represent the number of filter in layer $r$, $w_n^r$ and $b_n^r$ is the $nth$ filter and bias of layer $r$ respectively, $v_m^{r-1}$ is the $mth$ output from previous layer $r-1$, $*$ denotes the convolution operation. The obtained feature map is then processed with a pooling layer by taking the mean or maximum feature activation over disjoint regions. By cascading the combination of convolutional layer and pooling layer, a multi-layer structure is built for feature description. Finally, the fully-connected layers, just like the layers in multi-layer neural network, are employed for classification. Given the training set $\{X_j\}_j$, the learning process of a CNN with $K$ convolutional layers, including the parameters of filters $\{W_i\}_{i=1}^K$, the biases $\{b_i\}_{i=1}^K$ and classification layers $U$, can be defined as an optimization task:

$$\min_{\{W_i\}_{i=1}^K, \{b_i\}_{i=1}^K, U} \sum_j \ell(h(X_j), f(X_j, \{W_i\}_{i=1}^K, \{b_i\}_{i=1}^K, U)) \quad (2)$$

where $\ell$ denotes the loss function to estimate the cost between true label $h(X)$ and real predicted label by CNN model $f(X, \{W_i\}_{i=1}^K, \{b_i\}_{i=1}^K, U)$.

In fault diagnosis applications, CNN has also gained increasing interests. In [17], the authors investigated different time-frequency methods to convert the 1D vibration signal into time-frequency images, which were then fed into CNN for bearing condition diagnosis. In [18], the authors developed a feature learning method with multiscale layers in CNN to diagnose bearing faults from wavelet packet energy images. More generally, 1D convolutional structure is employed since it naturally fits the characteristics of the time-series signal. The works in [19] utilized 1D CNN to adaptively learn sensitive features from raw frequency spectrum of the data and achieved high diagnosis rates for combined gear-bearing-shaft faults in gearbox. The works in [20] applied the 1D CNN with wide kernels in the first convolutional layer to capture the low frequency features and restrain the high frequency noise from bearing fault signal. Although satisfactory diagnostic results can be observed in the respective case study, most of existing fault diagnosis applications of CNN still remain in the second stage, as discussed above. The explorations on the CNN based transfer learning is of great significance for fault diagnosis in industry application.

*B. Transfer learning*

Transfer learning is novel machine learning framework. For completeness, the definitions of transfer learning are first presented.

**Definition 1 (Domain)** A domain $\mathcal{D}$ is composed of two components: a feature space $\mathcal{X}$ and a marginal probability distribution $P(X)$, where $X = \{x_1, ..., x_n\} \in \mathcal{X}$ is a particular training dataset, i.e., $\mathcal{D} = \{\mathcal{X}, P(X)\}$.

**Definition 2 (Task)** A task $\mathcal{T}$ consists of two parts, a label space $\mathcal{Y}$ and a predictive function $f(X)$, which can be learned from the instance set $X$, i.e., $\mathcal{T} = \{\mathcal{Y}, f(X)\}$. Also, $f(X) = Q(Y|X)$ is the conditional probability distribution.

**Definition 3 (Transfer learning)** Given a source domain $\mathcal{D}_s$ with a learning task $\mathcal{T}_s$ and a target domain $\mathcal{D}_t$ with a learning task $\mathcal{T}_t$, transfer learning aims to facilitate the learning process of target predictive function $f_t(X)$ in $\mathcal{D}_t$ by using the related information or knowledge in $\mathcal{D}_s$ and $\mathcal{T}_s$, where $\mathcal{D}_s \neq \mathcal{D}_t$, or $\mathcal{T}_s \neq \mathcal{T}_t$. When the $\mathcal{D}_s = \mathcal{D}_t$ and $\mathcal{T}_s = \mathcal{T}_t$, the learning process can be identified with the traditional machine learning problem.

Two remarks should be emphasized here. The condition $\mathcal{D}_s \neq \mathcal{D}_t$ means $\mathcal{X}_s \neq \mathcal{X}_t \vee P_s(X_s) \neq P_t(X_t)$. And the condition of $\mathcal{T}_s \neq \mathcal{T}_t$ implies $\mathcal{Y}_s \neq \mathcal{Y}_t \vee Q_s(Y_s|X_s) \neq Q_t(Y_t|X_t)$.

In recent years, the broad application prospect of transfer learning has been viewed in different research areas. Several comprehensive surveys were made to review the present development of transfer learning [29]-[30]. In the surveys, transfer learning technology is categorized into many branches, such as inductive transfer learning, transductive transfer learning and unsupervised transfer learning. Domain adaptation, as a transductive transfer learning, fits the situation where the source domain labels are available, the target domain labels are unavailable, $\mathcal{X}_s = \mathcal{X}_t$ and $P_s(X_s) \neq P_t(X_t)$. As this situation is normally seen in practical problems for various fields, the domain adaptation has also received wide attention. Some algorithms concerned, such as transfer joint matching (TJM) [31], transfer component analysis (TCA) [32], joint distribution adaptation (JDA) [33] and deep transfer learning [34]-[36] have been designed gradually for image classification. In intelligent fault diagnosis, there are only a few works considering the application of transfer learning to strengthen the applicability and scalability of diagnosis framework for diverse domain tasks. In [37], the authors developed a SAE based domain adaptation method for bearing diagnosis across diverse operating conditions, where a maximum mean discrepancy (MMD) term is utilized to measure the domain discrepancy. In [38], the authors proposed a deep transfer network based domain adaptation method for bearing and gearbox diagnosis across diverse operating conditions. Specially, the model employed a MMD term to evaluate the discrepancy of normal category between source and target domains, and retained the sophisticated fault features with a weight regularization term. These studies have preliminarily explored the effectiveness of transfer learning in the field of intelligent fault diagnosis, but further works are needed to improve this framework in the following two aspects. 1) The transfer scenario should be extended to more challenging diagnosis tasks. The diverse fault severity levels and diverse fault types across domains also tend to result in large distribution discrepancy, where the attempts in these scenarios is necessary. 2) The previous studies only adapted the marginal distribution without considering the conditional distribution, leading to the neglect of the discrimination structures in rich labelled source data. Jointly reducing the discrepancy in both marginal distribution and conditional distribution may hold the potential to achieve superior transfer performance.

*C. Maximum mean discrepancy*

MMD is an index to measure the discrepancy of two distributions. Given two dataset $X_s, X_t$, $P_s(X_s) \neq P_t(X_t)$ and a nonlinear mapping function $\phi$ in a reproducing Kernel Hilbert space $\mathcal{H}$ (RKHS), the formulation of MMD can be defined as:

$$MMD_{\mathcal{H}}(X_s, X_t) = \left\| \frac{1}{n_s} \sum_{i=1}^{n_s} \phi(x_i^s) - \frac{1}{n_t} \sum_{i=1}^{n_t} \phi(x_i^t) \right\|_{\mathcal{H}} \quad (3)$$

where $n_s$ and $n_t$ are the numbers of samples in the two datasets respectively. In (3), we can find that the empirical estimation of the discrepancy for two distributions is considered as the distance between the two data distributions in RKHS. A value near zero for MMD means the two distributions are matched. In transfer learning, MMD is generally used to construct the regularization term for the constraint in feature learning, making the learned feature distributions more similar between different domains. In the neural network based transfer learning methods, MMD term is often appended to the loss function for optimization.

## III. DEEP TRANSFER NETWORK WITH JOINT DISTRIBUTION ADAPTATION

*A. Joint distribution adaptation*

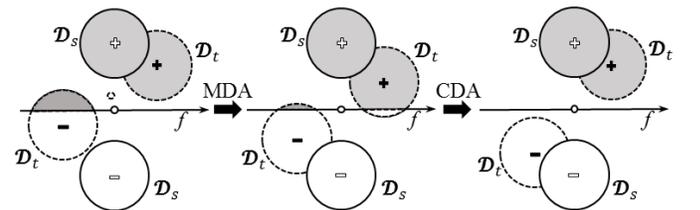

Fig. 2. An illustration of MDA and CDA, f: discriminative hyperplane, $\mathcal{D}_s$: feature distribution in source domain, $\mathcal{D}_t$: feature distribution in target domain.

Generally, the probability distributions of diverse domains may exhibit significant difference not only in marginal distribution, but also in conditional distribution for large amount of practical applications. Only adapting the marginal distributions, which represent the cluster structure of unlabeled data, cannot meet the required transfer performance, since the discriminative hyperplanes may also be different for diverse domain tasks. An intuitive description of this challenge is illustrated in Fig. 2. The marginal distribution adaptation (MDA) contributes to improving transfer performance, while the conditional distribution adaptation (CDA), which aims to match the discrimination structures between labeled source data and unlabeled target data, is also indispensable and highly effective. Hence, in this part, we are dedicated to presenting a simple mathematical formulation of JDA, and further providing a specific deep transfer framework.

**Problem Formulation (Joint distribution adaptation)** In a fault diagnosis task, given a labeled source dataset $X_s = \{x_i^s, y_i^s\}_{i=1}^{n_s}$ and a unlabeled target dataset $X_t = \{x_i^t\}_{i=1}^{n_t}$, $\mathcal{X}_s = \mathcal{X}_t$, $\mathcal{Y}_s = \mathcal{Y}_t$, $P_s(X_s) \neq P_t(X_t)$, $Q_s(Y_s|X_s) \neq Q_t(Y_t|X_t)$. The weak form of transfer learning with domain adaptation is to learn a feature transform that simultaneously minimizes the discrepancy between marginal distribution and conditional distribution [32], i.e.,

$$\min D(P_s(\phi(X_s)), P_t(\phi(X_t))) \quad (4)$$
$$\text{and } \min D(Q_s(Y_s|\phi(X_s)), Q_t(Y_t|\phi(X_t))) \quad (5)$$

where $D$ is the function to evaluate the domain discrepancy.

**1) MDA:** The objective function of (4) is to minimize the distance between the two data distributions in RKHS, where we can apply MMD (3) to tackle it. The formula is described as:

$$MMD_{\mathcal{H}}^2(P_s, P_t) = \left\| \frac{1}{n_s}\sum_{i=1}^{n_s}\phi(x_i^s) - \frac{1}{n_t}\sum_{i=1}^{n_t}\phi(x_i^t) \right\|_{\mathcal{H}}^2 \quad (6)$$

where $\phi: \mathcal{X} \to \mathcal{H}$ is the nonlinear mapping function in RKHS.

**2) CDA:** The conditional distribution in (5) is intractable in the absence of classification ground truth. We rewrite it into the following form:

$$\min D\left(\frac{Q_s(\phi(X_s)|Y_s) \cdot P_s(\phi(X_s))}{P(Y_s)}, \frac{Q_t(\phi(X_t)|Y_t) \cdot P_t(\phi(X_t))}{P(Y_t)}\right). \quad (7)$$

If the marginal distribution for (4) holds, the optimization problem in (7) becomes

$$\min D(Q_s(\phi(X_s)|Y_s), Q_t(\phi(X_t)|Y_t)). \quad (8)$$

The above objective function is noted as CDA. This step is essential for an accurate and robust distribution adaptation. However, it is still intractable as $Y_t$ is unknown. Some previous studies proposed a circuitous way by exploiting the pseudo labels for target data to handle the CDA in unsupervised domain adaptation. With the aid of the pre-trained models on labeled source data, pseudo labels for target data can be preliminarily supplied. Supposing a total of $C$ categories and the category $c \in \{1, \dots, C\}$, the distance index, MMD, can be defined to measure the mismatch of conditional distributions $Q_s(x_s|y_s = c)$ and $Q_t(x_t|y_t = c)$ of $c$ category,

$$MMD_{\mathcal{H}}^2(Q_s^{(c)}, Q_t^{(c)}) = \left\| \frac{1}{n_s^{(c)}}\sum_{x_i^s \in \mathcal{D}_s^{(c)}}\phi(x_i^s) - \frac{1}{n_t^{(c)}}\sum_{x_j^t \in \mathcal{D}_t^{(c)}}\phi(x_j^t) \right\|_{\mathcal{H}}^2 \quad (9)$$

where $\mathcal{D}_s^{(c)} = \{x_i : x_i \in \mathcal{D}_s \land y(x_i) = c\}$, $y(x_i)$ is the true label, and $n_s^{(c)} = |\mathcal{D}_s^{(c)}|$, $\mathcal{D}_t^{(c)} = \{x_j : x_j \in \mathcal{D}_t \land \hat{y}(x_j) = c\}$, $\hat{y}(x_j)$ is the pseudo label and $n_t^{(c)} = |\mathcal{D}_t^{(c)}|$.

It should be noted that, although there are probably many mistakes in the initial pseudo labels, one can iteratively update the pseudo labels in the stage of model optimization to obtain the optimal prediction accuracy under the current learning conditions.

**Joint distribution adaptation** (JDA): By integrating marginal MMD and conditional MMD, a regularization term of JDA can be written as:

$$D_{\mathcal{H}}(J_s, J_t) = MMD_{\mathcal{H}}^2(P_s, P_t) + \sum_{c=1}^{C} MMD_{\mathcal{H}}^2\left(Q_s^{(c)}, Q_t^{(c)}\right) \quad (10)$$

where $J_s$ and $J_t$ is the joint probability distribution of $\mathcal{D}_s$ and $\mathcal{D}_t$, respectively. Minimizing the (10) can guarantee the match both in marginal distribution and marginal distribution with sufficient statistics.

### B. Deep transfer network

Having introduced the regularization term of JDA, we now turn to the establishment of DTN, attempting to realize the goal

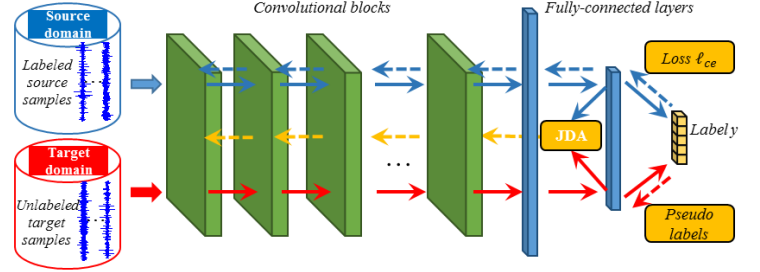

Fig. 3. The architecture of DTN for unsupervised domain adaptation.

of domain adaptation under deep learning framework. CNN is selected as the base model in this work.

Generally, we can train a CNN model on the sufficient source data from scratch with the optimization task defined in (2). The cross-entropy $\ell_{ce}$ between estimated probability distribution and true label is served as the loss function. When applying the pre-trained CNN model to domain adaptation, a new objective function is redefined by integrating the $\ell_{ce}$ and regularization term of JDA, rewritten as:

$$\mathcal{L}(\Theta) = \ell_{ce} + \lambda D_{\mathcal{H}}(J_s, J_t) \quad (11)$$

where $\Theta = \{W^i, b^i\}_{i=1}^{l}$ is the parameter collection of a CNN with $l$ layers and $\lambda$ is non-negative regularization parameter. It should be emphasized that the mapping function $\phi$ in RKHS $\mathcal{H}$ is the nonlinear feature transform learned by deep models herein. For CNNs, the features always change from general to specific with the increase of layer depth. The upper layers tend to represent more abstract features, which may result in a larger domain discrepancy [39]. Consequently, we deploy the regularization term on the last hidden fully-connected layer, namely the layer in front of discrimination layer, that is, $\phi(x) = h_{l-1}(x)$, where $h_{l-1}(\cdot)$ is the feature map by the nonlinear feature transform of the first $(l-1)$ layers. The JDA regularization term employed in conjunction with deep models can generate the mapping function $\phi$ by adaptively learning from data, and avoid to manually set the parameterized kernel function.

By minimizing the (11), the pre-trained CNN model can be further transferred and adapted for target tasks. The mini-batch stochastic gradient descent (SGD) and backpropagation algorithm are used for network optimization. The gradient of objective function with respect to network parameters is

$$\nabla_{\Theta^l} = \frac{\partial \ell_{ce}}{\partial \Theta^l} + \lambda \left(\nabla D_{\mathcal{H}}(J_s, J_t)\right)^T \left(\frac{\partial \phi(x)}{\partial \Theta^l}\right), \quad (12)$$

where $\frac{\partial \phi(x)}{\partial \Theta^l}$ are the partial derivatives of the output of $(l-1)$th layer with network parameters. The detailed formulations of $\nabla D_{\mathcal{H}}^2(J_s, J_t)$ are described as:

$$\nabla D_{\mathcal{H}}^2(J_s, J_t) = \nabla MMD_{\mathcal{H}}^2(P_s, P_t) + \sum_{c=1}^{C} \nabla MMD_{\mathcal{H}}^2\left(Q_s^{(c)}, Q_t^{(c)}\right), \quad (13)$$

$$\nabla MMD_{\mathcal{H}}^2(P_s, P_t) = \begin{cases} \frac{2}{n_s}\left(\frac{1}{n_s}\sum_{i=1}^{n_s}\phi(x_i^s) - \frac{1}{n_t}\sum_{j=1}^{n_t}\phi(x_j^t)\right) & x \in \mathcal{D}_s \\ \frac{2}{n_t}\left(\frac{1}{n_t}\sum_{j=1}^{n_t}\phi(x_j^t) - \frac{1}{n_s}\sum_{i=1}^{n_s}\phi(x_i^s)\right) & x \in \mathcal{D}_t \end{cases}, \quad (14)$$

$$\text{and } \nabla MMD_{\mathcal{H}}^2(Q_s^{(c)}, Q_t^{(c)}) = \begin{cases} \frac{2}{n_s^{(c)}}\left(\frac{1}{n_s^{(c)}}\sum_{x_i^s \in \mathcal{D}_s^{(c)}}\phi(x_i^s) - \frac{1}{n_t^{(c)}}\sum_{x_j^t \in \mathcal{D}_t^{(c)}}\phi(x_j^t)\right) & x \in \mathcal{D}_s \\ \frac{2}{n_t^{(c)}}\left(\frac{1}{n_t^{(c)}}\sum_{x_j^t \in \mathcal{D}_t^{(c)}}\phi(x_j^t) - \frac{1}{n_s^{(c)}}\sum_{x_i^s \in \mathcal{D}_s^{(c)}}\phi(x_i^s)\right) & x \in \mathcal{D}_t \end{cases}. \quad (15)$$

The architecture of proposed DTN with JDA is illustrated in Fig. 3.

### C. Training Strategy

---

**Algorithm 1** Training Procedure of DTN with JDA

**Input:** Given the dataset $D_s = \{x_i^s, y_i^s\}_{i=1}^{n_s}$ in source domain, unlabelled dataset $D_t = \{x_i^t\}_{i=1}^{n_t}$ in target domain, the architecture of deep neural network, the trade-off parameters $\lambda$
**Output:** Transferred network and predicted labels for target samples
1: **begin**
2: Train a base deep network on the source dataset $D_s$
3: Predict the pseudo labels $\hat{Y}_0 = \{y_i^t\}_{i=1}^{n_t}$ for target samples with base network
4: **repeat**
5:     $j = j + 1$
6:     Compute the regularization term of JDA according to (10)
7:     Network optimization with respect to (11)
8:     Update the pseudo labels $\hat{Y}_j$ with optimized network
9: **until** convergence or $\hat{Y}_j = \hat{Y}_{j-1}$
10: Check the diagnosis performance of transferred network on other target samples.

---

The training procedure of this framework mainly consists of two parts: 1) the pre-training on labeled source data and 2) the network adaptation in target domain with the input of both labelled source data and unlabeled target data. It should be noted that the dataset is generally divided into small batches, which are fed into the network for training. A desirable batch size should be as large as possible to cover the variance of the whole dataset, whereas a too large batch size will also increase the calculation burden. It is a trade-off between transfer performance and computational effectiveness. Besides, the same amount of samples from source and target domain are used for network adaptation. When the data sizes are different across domains, the re-sampling can be applied in the smaller dataset to keep the same number of samples in the source and target datasets. The complete training procedure of DTN with JDA is summarized in Algorithm 1.

### D. Comparison studies

1) Comparison methods: The proposed framework will be compared with several state-of-the-art methods in the field of intelligent fault diagnosis: 1) SVM [10]; 2) Random forest (RF) [10]; 3) Empirical mode decomposition analysis (EMD) [1]; 4) CNN; 5) TJM [31]; 6) TCA [32]; 7) JDA [33]; 8) DTN with MDA and 9) DTN with JDA (this work). These baseline methods can be categorized into two subsettings: the standard diagnosis methods 1)-4) and the transfer learning based techniques 5)-9).

In 1)-2), the 29 popular statistical features [7] are extracted from raw data in time and frequency domains to form the input of the classifiers. In 3), EMD is used to decompose the original signal into a series of intrinsic mode functions (IMF). The energy distribution of first five IMFs is calculated as the input features for classifier. In 4), using the deep learning flow, CNN.

In the transfer learning based techniques 5)-9), TJM, TCA and JDA are the shallow transfer learning methods, and thus we also extract the 29 statistical features from raw data, then conduct the unsupervised domain adaptation, finally make diagnosis results with classifier. In deep learning flow, a comparison between the proposed method and DTN with MDA method by removing the CDA term in objective function is made. The pre-trained base network resorts to the optimal CNN model in source domain, that is, the trained model in 4).

2) Implementation details: For 1)-4), we use the labeled source data to train the model, which will be applied to diagnose unlabeled target data. For 5)-7), TJM, TCA and JDA simultaneously process the labeled source data and unlabeled target data for dimension reduction. The classification model is then trained with low-dimensional features in source domain, and deployed on target domain. Herein, both the SVM and RF are adopted to achieve an optimal diagnosis performance as the final results. The RBF is adopted as the kernel in SVM, and the penalty factor and kernel function parameter are both empirically set to 8 for further discussion. The two structure parameters in RF, i.e., number of trees $n_{try}$ and the number of random feature subset $m_{try}$ are set to 500 and $\lfloor\sqrt{m}\rfloor$ respectively, where $m$ is the dimension of input vector. Referring to literatures of TJM, TCA and JDA, the adaptation regularization parameter $\lambda$ is set by searching $\lambda \in \{1e^{-2}, 1e^{-1}, 1, 10, 20, 50, 100\}$. The kernel type is selected from RBF and linear, and the dimension after adaptation is optimized with the strategy of trial and error. In DTN methods, the adaptation regularization parameter $\lambda$ is set by searching $\lambda \in \{1e^{-4}, 1e^{-3}, 1e^{-2}, 5e^{-2}, 1e^{-1}, 5e^{-1}, 1\}$. The learning rate of SGD is set to 0.01.

## IV. EXPERIMENTS

In this section, experiments on three mechanical fault datasets are conducted to demonstrate the efficiency, superiority as well as practical value of proposed transfer framework.

### A. Data description

1) Wind Turbine Fault Dataset: The first dataset is from our wind turbine experimental platform. This dataset contains ten machine conditions, which are health, front bearing pedestal loosening (FB), back bearing pedestal loosening (BB), rolling element fault of front bearing (RF), inner-ring fault of front bearing (IF), outer-ring fault of front bearing (OF), misalignment in horizontal direction (MH), misalignment in vertical direction (MV), variation in airfoil of blades (VB) and yaw fault (YF) respectively (corresponding labels 0-9). All these faults can basically simulate the typical failure modes from wind wheel to drive chain of a real wind turbine.

To create working conditions close to reality, we change the power of axial flow fan in the wind tunnel to generate varying loading conditions (i.e., varying wind speeds). The experiments are performed under six different wind speeds ranging from 5.8m/s to 11.5m/s (loads 0-5). And the corresponding speeds of wind wheel range from 255rpm to 300rpm. The vibration data is collected by two accelerometers, which are installed on front and back bearing pedestal respectively, with a sampling frequency of 20 kHz. The signal segments from the two accelerometers are combined together as a sample for fusing the data. Detailed descriptions of experimental scheme can be referred in [40].

For clarity, the denotation of A→B is utilized to represent the transfer task from source dataset A to target dataset B. In the wind turbine fault dataset, we aim to explore the transfer ability of proposed framework across diverse operating conditions. Consequently, six transfer tasks are designed for empirical evaluation: A→B, B→A, C→D, D→C, E→F and F→E. For instance, A→B: the source dataset A contains the samples of ten machine conditions under load 0-2, while the target dataset B is composed of the samples under load 3-5. Details of tasks design are listed in Table I.

TABLE I
DESIGNED TRANSFER TASKS ACROSS DIVERSE OPERATING CONDITIONS

| Transfer tasks | Source domain | Target domain | Unlabelled target samples | Testing target sample | Machine conditions |
|---|---|---|---|---|---|
| A→B | Load 0-2 | Load 3-5 | 24000 | 4000 | 10 conditions (labels 0-9) |
| B→A | Load 3-5 | Load 0-2 | 24000 | 4000 | |
| C→D | Load 2 | Load 3-5 | 24000 | 4000 | |
| D→C | Load 3-5 | Load 2 | 12000 | 4000 | |
| E→F | Load 2 | Load 5 | 12000 | 4000 | |
| F→E | Load 5 | Load 2 | 12000 | 4000 | |

2) *Bearing Fault Dataset:* The bearing fault dataset is an open-access dataset from Case Western Reserve University [41]. Four different bearing conditions, i.e., health, outer race fault (OF), roller fault (RF) and inner race fault (IF) (corresponding labels 0-3), are considered in this dataset. The experiments are performed under four motor speeds (1797rpm, 1772rpm, 1750rpm and 1730rpm) at a sampling frequency of 12 kHz. For each fault type, single point faults with different severity levels are introduced to test bearing respectively. In most existing studies, the samples with the same fault type but different severity levels are treated as distinct categories. Indeed, the signal characteristic of certain fault type always varies with the severity level, and it will lead to a large domain discrepancy in real industry diagnosis application when target conditions are not well represented the source scope whose sufficient data is used to train the diagnosis model. Therefore, we aim to investigate the performance of proposed transfer framework across diverse fault severities in this dataset.

For simplicity, we select two fault severity levels with the fault diameters (FD) of 0.18mm and 0.53mm to construct transfer tasks: G→H, H→G. The dataset G is composed of the samples of four bearing conditions under four motor speeds and the fault diameter of OF, RF and IF cases is 0.18mm. The dataset H is formed by the health samples and fault samples with 0.53mm fault diameter.

3) *Gearbox Fault Dataset:* The gearbox fault dataset collected from our single-stage cylindrical straight gearbox test rig is analyzed in the scenario where the domain discrepancy between specific fault types are expected to be bridged by transfer learning. Sometimes, it may be more practical to confirm the location of failure instead of specific types. Considering the example of gearbox, identifying the fault location, such as gear fault or bearing fault, is beneficial for monitoring and maintenance. That said, certain types of fault occurred in one component, such as bearing inner race fault or outer race fault, can be defined as one category. Besides, it may be impossible to obtain the fault data of various fault types and train a diagnosis model with high accuracy for a complex mechanical system. Consequently, the transfer performance across similar but diverse fault types is of great practical significance. In the experiments, we introduced two types of faults, i.e., gear root crack (RC) and tooth surface spalling (TS), to high-speed cylindrical gearing, and another two types of faults, i.e., outer race fault (OF) and roller fault (RF), to high-speed conical bearing. The vibration data is collected under two working speeds (900rpm and 1500rpm). The sampling frequency is 20 kHz.

We state three conditions of gearbox, including health, gear fault and bearing fault in this dataset (corresponding labels 0-2), and design two transfer tasks: I→J, J→I. The dataset I contains the samples of health, bearing OF and gear RC. The dataset J is formed by the samples of health, bearing RF and gear TS.

TABLE II
DESIGNED TRANSFER TASKS ACROSS DIVERSE FAULT SEVERITY LEVELS AND TYPES

| Transfer tasks | Source domain | Target domain | Unlabelled target samples | Testing target sample | Machine conditions |
|---|---|---|---|---|---|
| G→H | FD 0.18 | FD 0.53 | 12000 | 4000 | 4 conditions (labels 0-3) |
| H→G | FD 0.53 | FD 0.18 | 12000 | 4000 | |
| I→J | Health, OF, RC | Health, RF, TS | 12000 | 4000 | 3 conditions (labels 0-2) |
| J→I | Health, RF, TS | Health, OF, RC | 12000 | 4000 | |

*B. Results*

The diagnosis results of ten tasks are shown in Table III and Fig. 4, respectively. Several encouraging observations are firstly noted. 1) The DTN with JDA method in this work significantly outperforms the other methods. The stable accuracies under different transfer scenarios (almost over 95% for all tasks) validate the effective and robust domain adaptation ability of proposed method. 2) The diagnosis performance in standard methods (the first four) is much improved with domain adaptation for most cases. Especially, the average accuracy of DTN with JDA is 97.5%, and makes a 17.5% transfer improvement, comparing with the baseline CNN, 80.0%. 3) The deep learning methods always present a superior performance to the shallow methods no matter in standard diagnosis framework or transfer learning framework, conforming its extraordinary feature learning and representation capacity as well as a stronger feature transferability. 4) By jointly adapting the marginal distribution and conditional distribution, the DTN with JDA in this work significantly promotes the adaptation ability of previous DTN with MDA, especially under the transfer scenarios of diverse fault severity levels and diverse fault types.

Secondly, in the standard methods, the diagnosis performance varies a lot on different tasks. For instance, in the first six tasks on wind turbine dataset, RF get the best performance in task C→D and D→C (88.8% and 92.8%), while degraded results in tasks E→F and F→E (77.3% and 60.8%). It is reasonable because the operating conditions between C and D are closer than those between E and F, and thus the data for C and D shares a more similar feature space, leading to a higher diagnosis accuracy. This phenomenon actually reveals the inherent drawback in conventional diagnosis framework. The success much relies on the similarity between source and target data, whereas a large discrepancy across domains is common and inevitable in practical diagnosis applications.

TABLE III
DIAGNOSIS ACCURACY (%) ON TEN TRANSFER TASKS WITH DIFFERENT METHODS

| Methods | A→B | B→A | C→D | D→C | E→F | F→E | G→H | H→G | I→J | J→I | Average |
|---|---|---|---|---|---|---|---|---|---|---|---|
| SVM | 77.9 | 75.4 | 79.8 | 90.7 | 63.1 | 62.5 | 73.3 | 76.2 | 78.5 | 46.8 | 72.4 |
| RF | 85.1 | 79.6 | 88.8 | 92.8 | 77.3 | 60.8 | 81.2 | 50.8 | 69.1 | 69.4 | 75.5 |
| EMD | 79.7 | 71.0 | 82.3 | 84.8 | 79.8 | 73.5 | 40.0 | 50.9 | 53.4 | 44.0 | 65.9 |
| CNN | 91.8 | 92.3 | 86.3 | 92.6 | 77.0 | 79.7 | 78.8 | 70.0 | 74.7 | 56.9 | 80.0 |
| TJM | 87.9 | 79.3 | 89.5 | 93.7 | 80.5 | 68.3 | 93.6 | 97.2 | 73.2 | 56.1 | 81.9 |
| TCA | 89.7 | 81.6 | 89.3 | 93.2 | 82.6 | 72.7 | 92.4 | 94.9 | 72.9 | 49.2 | 81.9 |
| JDA | 85.9 | 80.9 | 92.9 | 93.8 | 86.1 | 65.9 | 92.1 | 99.2 | 76.6 | 47.7 | 82.1 |
| DTN with MDA | 95.6 | 96.7 | 94.8 | 98.5 | 87.0 | 88.4 | 87.2 | 63.7 | 84.0 | 78.1 | 87.4 |
| DTN with JDA | **97.9** | **98.2** | **96.4** | **98.8** | **95.8** | **94.3** | **98.1** | **100.0** | **99.7** | **96.1** | **97.5** |

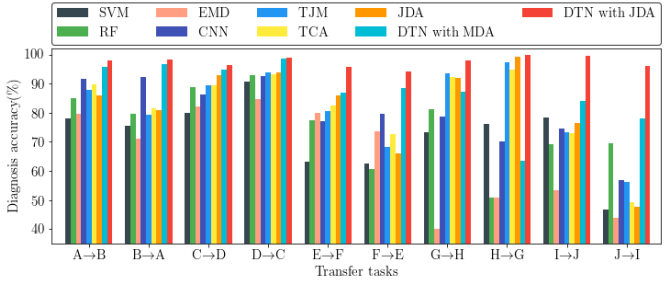

Fig. 4. Comparison of the performance of diverse methods on ten transfer tasks.

Thirdly, it is worth noting that the complexity of the domain adaptation process always changes with the scenarios. In the easy transfer tasks, e.g. C→D and D→C, all these transfer learning based techniques get the relatively satisfactory results. However, in several hard tasks, e.g. E→F and J→I, where the source and target data could be substantially dissimilar, the performance drop in the comparative transfer methods, such as DTN with MDA, convincingly illustrate that the difficulties of domain adaptation will accordingly increase. The comprehensive assessments under diverse transfer scenarios further demonstrate the pivotal role of JDA in DTN.

### C. Network visualization: how JDA outperforms MDA

In order to give a clear and intuitive understanding of proposed framework, a nonlinear dimensionality reduction, namely, t-distributed stochastic neighbor embedding (t-SNE), is employed for network visualization. For comparison, the visualization results of standard CNN (that is, the pre-trained base network for further domain adaptation), DTN with MDA and DTN with JDA in three transfer tasks are presented in Fig. 5-7, respectively.

Task E→F is to realize the domain adaptation across diverse operating conditions. First, as shown in Fig. 5(a), most of the 10 categories of source samples are well separated with the standard CNN, while the feature distributions of same category between source and target domains are not aligned well. And even worse, a large overlapping areas can be inspected among the target samples of certain categories, such as 2, 3 and 8. These observations suggest the domain discrepancy exists not only in marginal distribution, but also in conditional distribution, which may result in the degraded diagnosis results in conventional framework. In Fig. 5(b) and (c), under the transfer learning framework, we can find the obvious improvement of distribution adaptation. In particular, the same category between domains is aligned very well by DTN with JDA, and a consonant and legible discriminant structure can be observed for both source and target categories.

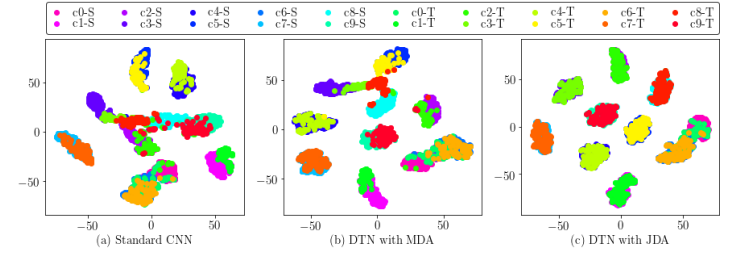

Fig. 5. Network visualization in task E→F: t-SNE is applied on the feature representation of last hidden fully-connected layer for both the source data and target data. There are total 10 categories in wind turbine dataset (corresponding labels 0-9). S represents the samples in source domain and T means the target domain. For instance, the c5-T corresponds to the samples of category 5 (inner-ring fault of bearing, as introduced above) in target domain.

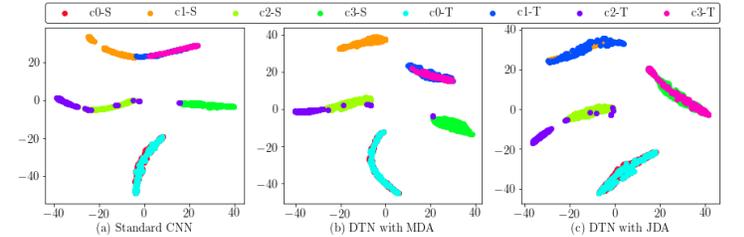

Fig. 6. Network visualization in task G→H: there are total 4 categories in bearing dataset (corresponding labels 0-3).

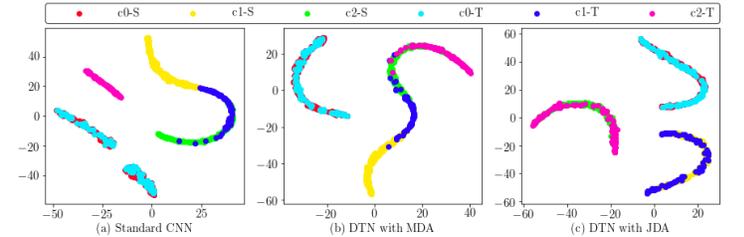

Fig. 7. Network visualization in task I→J: there are total 3 categories in gearbox dataset (corresponding labels 0-2).

Task G→H is to adapt the distribution across diverse fault severity levels. In Fig. 6(a), the standard CNN assembles the distributions of OF and IF in target domain, and the source OF and target IF are mixed, explaining the unsatisfactory accuracy in Table III. As a contrast, in Fig. 6(c), the distribution of same category between source and target domains are well matched with JDA. Interestingly, in Fig. 6(b), we can observe that MDA relocates the target OF and IF away from the corresponding distribution in source domain. Naturally, marginal distribution only reflects the cluster structure for the feature distribution of all categories, and MDA aims to explicitly reduce the distance between the cluster centers of different domains. When the conditional distributions are same across domains, MDA helps to correct the overall shift of feature space. However, in the field of fault diagnosis, the difference in conditional distributions may be prevalent. Consequently, unlike single MDA, the JDA which simultaneously adapts the marginal distribution and conditional distribution is promising in these cases. As shown in Fig. 7, similar results can be found in transfer task I→J across diverse fault types. Both the transfer accuracy and network visualization show that JDA supersedes the performance of MDA.

*D. Why JDA is better than MDA*

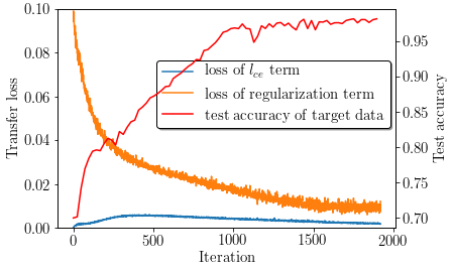

Fig. 8. The transfer loss curves and test accuracy via DTN with JDA.

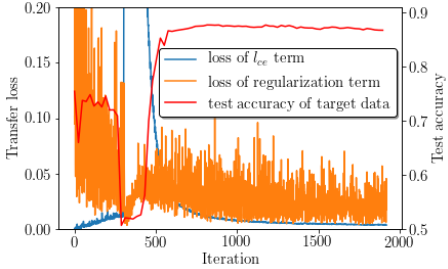

Fig. 9. The transfer loss curves and test accuracy via DTN with MDA.

**1) JDA converges smoothly with better accuracy:** Since the additional regularization term is appended to objective function for the transfer training, the convergence analysis is necessary to illustrate the transfer ability. The transfer loss curves and test accuracy curve for DTN with JDA and DTN with MDA in task G→H are plot in Fig. 8 and 9, respectively. Here, we separately display the $\ell_{ce}$ term and regularization term for ease of observation. At the beginning, the losses of regularization term for the two methods are both around 0.1, and the ones of $\ell_{ce}$ term are almost negligible. From Fig. 8, the loss of JDA regularization term converges to a certain degree after a series of iterations, accompanied by the continued increase of test accuracy of target data. However, from Fig. 9, the loss of MDA regularization term finally fluctuates in a high level, and the test accuracy is confined around 87%. Besides, it is clear to observe the loss of $\ell_{ce}$ term presents an abrupt increase after around 300 iterations. Essentially, the $\ell_{ce}$ term and regularization term in objective function try to reduce domain discrepancy while preserving the original discriminant structure in source domain. One possible reason for the jump is that the gradient direction of the parameter optimization for regularization term conflicts with that of $\ell_{ce}$ term, causing a significant spike in transfer loss and test accuracy. The analysis reveals that the use of JDA regularization term is capable of facilitating the network training and guaranteeing a stronger feature transferability.

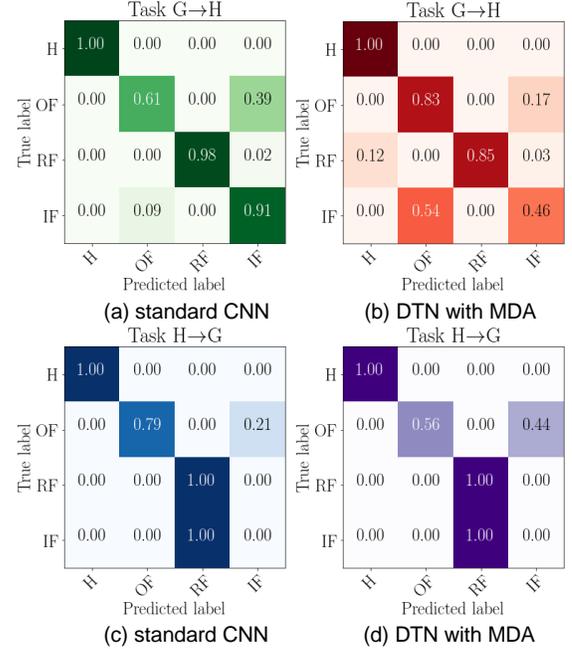

Fig. 10. Normalized confusion matrixes in task G→H and task H→G.

**2) JDA avoids negative adaptation:** Another remarkable point is the transfer result of DTN with MDA (63.7%) is lower than the result without transfer (70%) in task H→G (Table III). The negative adaptation probably occurs when the distribution of target domain is largely unlike that of the source domain, especially in the scenario that some of the conditions are with more significant variation between the source and target domains. As illustrated in Fig. 10, in task G→H, the MDA has a positive influence on domain adaptation, whereas task H→G presents the opposite case. We can find the IF samples are completely misclassified into RF by the pre-trained network (panel (c)), which can get high diagnosis accuracy for source samples. This will lead to significant distribution difference between domains in the initial stages of domain adaptation, undoubtedly increasing the transfer difficulty. Fortunately, by employing the CDA, the DTN with JDA can avoid this kind of negative adaptation and achieves high accuracy (100% in Table III) in this situation.

## V. CONCLUSION

Intelligent fault diagnosis in real industrial applications is suffering the difficulty of model re-training as of the discrepancy between the source domain (where the model is learnt) and the target domain (where the model is applied). However, re-training the model is challenging and probably unrealistic as of the lack of sufficient labeled data in practical applications. To address this issue, this work presents a DTN to take advantage of a pre-trained network from the source domain and get the model transferred with unlabeled data from the target domain, where a novel domain adaptation approach, JDA, is presented. Compared with the MDA, the necessity of the CDA in fault diagnosis is illustrated and the JDA is properly formulated within the DTN. Through extensive experiments on three datasets, the results show that the DTN with JDA outperforms the state-of-the-art approaches and its superiority over MDA is validated with network visualization. Furthermore, the DTN with JDA presents smooth convergence and avoids negative adaptation in comparison with MDA.

Using DTN with JDA, it is promising that the learnt diagnosis models from experimental or real datasets can be transferred to new but similar applications in a more efficient and accurate way, which could benefit kinds of industrial applications. Further work will pursue (i) more case studies on the real data, and (ii) application on imbalanced distribution of machine conditions.